# Self Driving RC Car using Behavioral Cloning

### [1.]Aliasgar Haji, [2.]Priyam Shah, [3.]Srinivas Bijoor

*[1]Student, IT Dept., K. J. Somaiya College Of Engineering, Mumbai.*
*[2]Student, Computer Dept., K. J. Somaiya College Of Engineering, Mumbai.*
*[3]Student, Startup School India, K. J. Somaiya College Of Engineering, Mumbai.*

-------------------------------------------------------------------***-------------------------------------------------------------------

**Abstract -** *Self Driving Car technology is a vehicle that guides itself without human conduction. The first truly autonomous cars appeared in the 1980s with projects funded by DARPA( Defense Advance Research Project Agency ). Since then a lot has changed with the improvements in the fields of Computer Vision and Machine Learning. We have used the concept of behavioral cloning to convert a normal RC model car into an autonomous car using Deep Learning technology.*

## 1.INTRODUCTION

Traveling by car is currently one of the most dangerous forms of transportation, with over a million deaths annually worldwide. As nearly all car crashes (particularly fatal ones) are caused by driver error, driverless cars would effectively eliminate nearly all hazards associated with driving as well as driver fatalities and injuries. A self-driving car is a vehicle equipped with an autopilot system and is capable of driving from one point to another without aid from a human operator. Self-driving car technology was built initially using the robotics approach. But with the advancement in the field of computer vision and machine learning, we can use the deep learning approach. Major contests are conducted in the US for self-driving car technology to make it available to the world. Some of the well-known projects are EUREKA Prometheus Project (1987-1995) ARGO Project, Italy (2001) DARPA Grand Challenge (2004-2007) European Land-Robot Trial (2006-2008). There are several challenges are needed to be met before implementing the self-driving car in the real world. It has to navigate through desert, flat and mountainous terrains and handle obstacles like bridges, underpasses, debris, potholes, pedestrians and other vehicles.

### 1.1 Neural Network

Neural networks are a set of algorithms, modeled loosely after the human brain, that is designed to recognize patterns. They can predict the output over a set of new input data through a kind of machine perception, labeling or clustering raw input as they learn to recognize and interpret pattern.

### 1.2 Convolutional Neural Networks (CNNs / ConvNets)

Convolutional Neural Networks as they are made up of neurons that have learnable weights and biases are similar to ordinary Neural Networks. [1] Each neuron receives some inputs, performs a dot product and optionally follows it with a non-linearity activation function. The overall functionality of the network is like having an image on one end and class as an output on the other end. They still have a loss function like Logarithmic loss/ Softmax on the last (fully-connected) layer and all ideas developed for learning regular Neural Networks still apply. In simple words, images are sent to the input side and the output side to classify them into classes based on a probability score, whether the input image applied is a cat or a dog or a rat and so on.

### 1.3 Behavioral Cloning

[2]Behavioral cloning is a method by which sub-cognitive skills like -recognizing objects, experience while performing an action can be captured and reproduced in a computer program. The skills performed by human actors are recorded along with the situation that gave rise to the action. The input to the learning program is the log file generated from the above data collection. The learning program outputs a set of rules that reproduce skilled behavior. The application of this method can be to construct automatic control systems for complex tasks for which classical control theory is inadequate.

## 2. Literature review

[3] NVIDIA has used convolutional neural networks (CNNs) to map the image from a front-facing camera to the steering commands for a self-driving car. The end-to-end approach is powerful because, with minimum training data from humans, the system learns to steer, with or without lane markings, on both local roads and highways.

The system is designed using an NVIDIA DevBox running Torch 7 for training. An NVIDIA DRIVE PX self-driving car computer, also with Torch 7, was used to determine where to drive—while operating at 30 frames per second (FPS). The automation in the training of the system to learn the representations of necessary processing steps, such as detecting useful surrounding features, with only the human steering angle as the training feature. NVIDIA never explicitly trained its system to detect, the outline of roads and other obstructions. It learned using the explicit decomposition of the problem, such as lane marking detection, path planning, and control, our end-to-end system optimizes all processing steps simultaneously.

It is believed that end-to-end learning leads to better performance over part by part learning models. Better performance is achieved due to the internal components which self-optimize to maximize overall system performance, instead of optimizing human-selected intermediate criteria, (e. g., lane detection). Using end-to-end learning can lead to smaller networks possible because the system learns to solve the problem with the minimal number of processing steps.

**Hardware Design**

Nvidia implemented Behavioral Cloning using its Drive PX Graphics Processing Unit. ( Fig - 1)

- The hardware consists of 3 cameras.

- The vehicle's Controller Area Network (CAN) bus provides steering commands.

- Nvidia's Drive PX onboard computer with GPUs.

- In order to make the system independent of the car geometry, the steering command is 1/r, where r is the turning radius in meters.

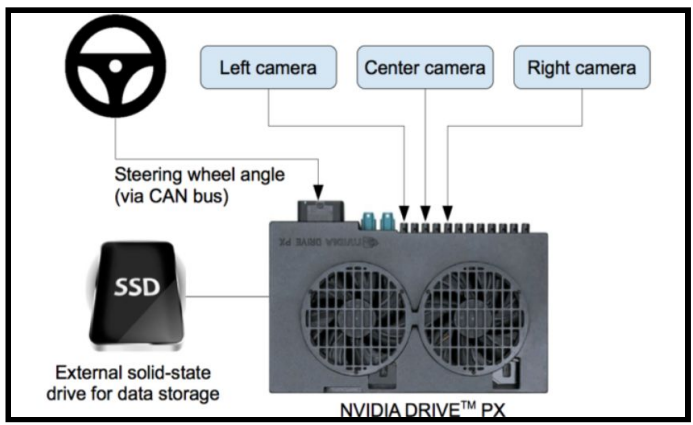

**Fig - 1: Hardware Design of NVIDIA Drive PX**

**Software Design**

(Fig - 2) The input to the CNN model is an image and output is steering command. The resultant command is compared to the required command for that image, and the weights of the CNN are adjusted to bring the output closer to the required output. The weight adjustment is accomplished using backpropagation. The network is able to generate steering commands from the video images of a single-center camera after being trained.

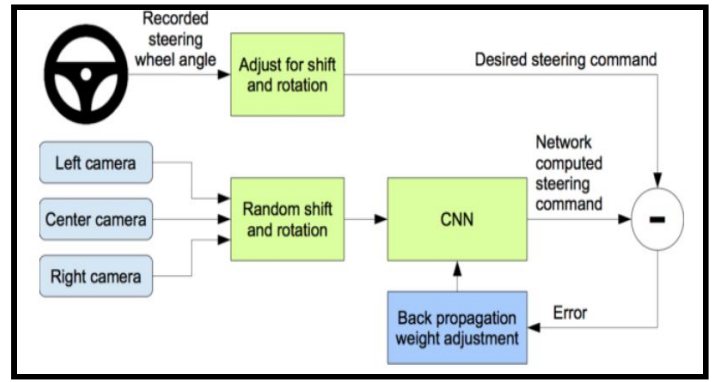

**Fig - 2: Software Design of NVIDIA Drive PX**

## 3. Our Implementation

(Fig-3) Our implementation comprises of Hardware stack which consists of the RC Car, Raspberry Pi model 3B, Pi Camera, jumper wires and the Software Stack consists of Raspbian OS, Keras, CV2 and gpio library of python.

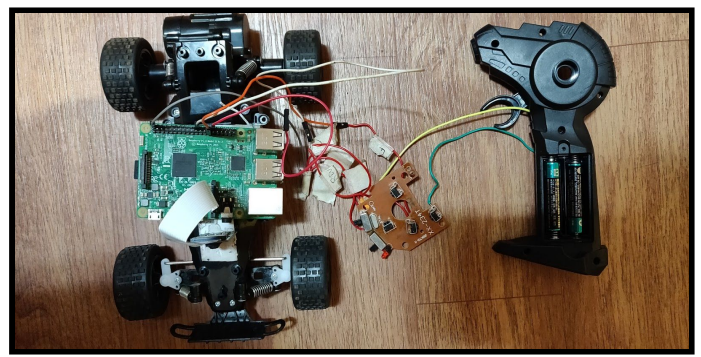

**Fig -3:  Our Implementation**

### 3.1 Hardware Implementation:

We bought a normal RC car, which is controlled wirelessly using the remote. We reverse-engineered its remote and soldered the-forward, reverse, right and left pins with male to female jumper wires such that it can be connected with the appropriate gpio pins of the Raspberry Pi. After this, we connected the Raspberry Pi with the PC and stored the weight file for prediction. The Pi Camera is placed on the car and the images are sent to the Raspberry Pi. The Pi then signals the next command-straight, right and left according to the prediction of the Deep Learning model.

### 3.2 Software Implementation:

The Software Implementation is divided into three main sections data generation, training phase, and testing phase.

### 3.2.1 Data Generation

The Pi Camera module records video feed and generates images of resolution 648 X 648. A .csv data file is generated along with the steering details (0:straight, 1-right and -1: left) with respect to images in each frame.

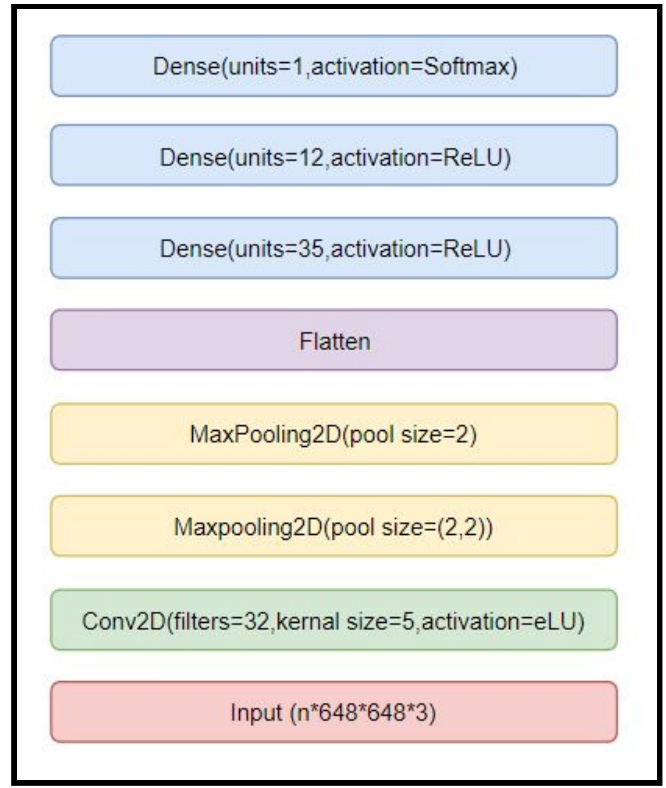

**Fig - 4: Generated Dataset**

### 3.2.2 Training phase

The video feed recorded by the camera was of resolution 648 X 648. For training, we kept the batch size to 32. We used CNN sequential layers for training (Fig - 5) and compiled the model file. Sequential layers are basically a linear stack of layers. The first layer added to sequential layer used is a two-dimensional convolution layer. We also used the activation function called "elu" to the first layer. ELU is very similar to RELU except for negative inputs. The second layer added was a MaxPooling layer. The pooling layer basically reduces computation by sending inputs with fewer parameters. The third layer was to flatten the inputs from previous layers. The fourth, fifth, sixth layers are dense layers. For the third and fourth layers we used ReLu activation function. For the fifth layer, Softmax activation function was used for multi-class classification. We split the dataset using pandas and sklearn libraries of python. The dataset split was of the ratio 80/20 where 80% is for training and 20% for testing. We also used a Adam optimizer during the training process and used "mean-square error" method to reduce the loss, as a loss function. We set epoch as 20 and a batch size of 8 for training.

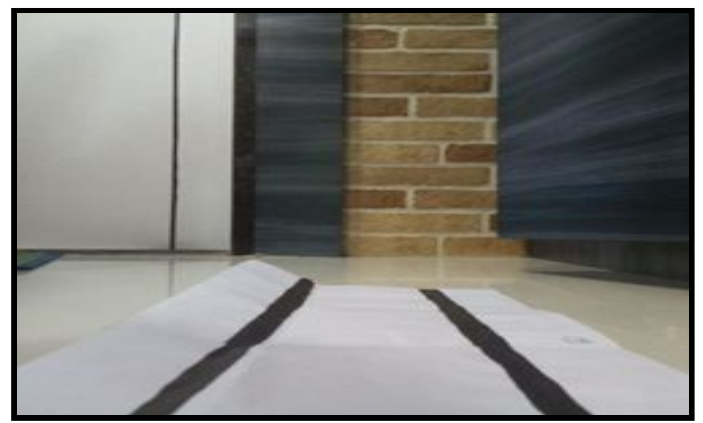

**Fig - 5: Architecture of Network**

### 3.2.3 Testing Phase

The model pickle file generated after the training step is used for testing, with real-time images and we got results as shown below.

**Fig - 6: Output: Forward**

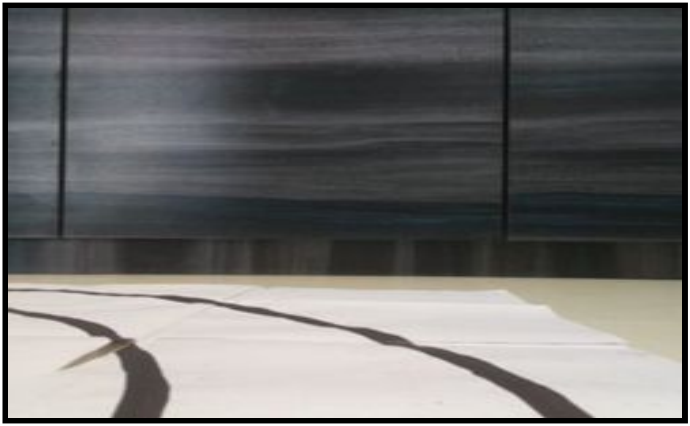

**Fig - 7: Output: Left**

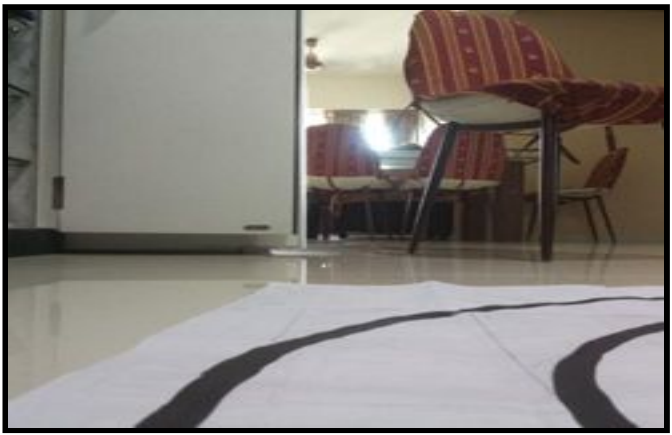

**Fig - 8: Output: Right**

## 4.Result and Observations

The main aim was to check how feasible it is to train a car using behavioral cloning so that it can drive autonomously in a given environment, instead of using other complex and computationally expensive methods like reinforcement learning. The model is trained over the labelled images captured while driving the RC car manually. The images clicked are mapped with the commands of- left, right or straight using the RPI-gpio package of the RaspberryPi to the remote and eventually to the RC car. This makes it possible for the car to make appropriate decisions when driven in the autonomous mode. We got a test accuracy of 84.5% after testing the Self Driving car in similar driving conditions.

[4]The main advantage of using behavioral cloning is that a lot of training data is not required. After driving the car for approximately thirty minutes and collecting about 200 images. We were successful in training our model to obtain a reasonable accuracy. Also, by using data augmentation one can increase the dataset and improve the model.

[5]However, this approach has two major disadvantages: one is that it needs an expert human driver so that the model gets trained properly. The other is that this approach cannot deal with dynamically changing scenarios. For example, if there are moving obstacles on the path or drastic change in the road structure, the model would fail. We can use DAgger (Dataset Aggregation algorithm) method for generalizing and solving these two problems.

## 5. Conclusion

Our car performed satisfactorily. But there was an observable lag due to the low-computing capability of Raspberry Pi model 3b. Hence, the steering decisions made by the car, although correct were not made in a suitable time frame. Thus, this method is suitable for training and implementing an autonomous car for a set environment, that does not change unpredictably.

## 6. Future Scope

To improve the current implementation we suggest using an advanced algorithm for Image processing.A GPU(Graphic Processing Unit) can be used for faster processing. Path navigation with GPS guidance Vehicle and Object detection features can be added. We can apply sensor fusion techniques, localization, and control features to improve the autonomous behavior of the vehicle.